\documentclass[10pt,twocolumn,letterpaper]{article}

\usepackage{cvpr}
\usepackage{times}
\usepackage{epsfig}
\usepackage{graphicx}
\usepackage{amsmath}
\usepackage{amssymb}
\usepackage{graphicx,stackengine,scalerel}

\usepackage{subfigure}
\usepackage{booktabs} %

\usepackage[pagebackref=true,breaklinks=true,letterpaper=true,colorlinks,bookmarks=false]{hyperref}

\cvprfinalcopy %

\def\cvprPaperID{1503} %

\graphicspath{{./}{./}}

\usepackage{multirow}
\newlength\savewidth\newcommand\shline{\noalign{\global\savewidth\arrayrulewidth
  \global\arrayrulewidth 1pt}\hline\noalign{\global\arrayrulewidth\savewidth}}

\newcommand{\first}[1]{\textcolor{red}{\textbf{#1}}}

\usepackage{pifont}%
\newcommand{\cmark}{\ding{51}}%

\ifcvprfinal\fi
\begin{document}

\title{SiamRPN++: Evolution of Siamese Visual Tracking with Very Deep Networks}

\author{Bo Li\thanks{The first three authors contributed equally. Work done at SenseTime. Project page: \url{https://lb1100.github.io/SiamRPN++}.}\\
SenseTime Research\\
{\tt\small libo@sensetime.com}
\and
Wei Wu$^*$\\
SenseTime Research\\
{\tt\small wuwei@sensetime.com}
\and
Qiang Wang$^*$\\
NLPR, CASIA\\
{\tt\small qiang.wang@nlpr.ia.ac.cn}
\and
Fangyi Zhang\\
VIPL, ICT\\
{\tt\small fangyi.zhang@vipl.ict.ac.cn}
\and
Junliang Xing\\
NLPR, CASIA\\
{\tt\small jlxing@nlpr.ia.ac.cn}
\and
Junjie Yan\\
SenseTime Research\\
{\tt\small yanjunjie@sensetime.com}
}

\maketitle

\begin{abstract}

Siamese network based trackers formulate tracking as convolutional feature cross-correlation between a target template and a search region. However, Siamese trackers still have an accuracy gap compared with state-of-the-art algorithms and they cannot take advantage of features from deep networks, such as ResNet-50 or deeper. In this work we prove the core reason comes from the lack of strict translation invariance. By comprehensive theoretical analysis and experimental validations, we break this restriction through a simple yet effective spatial aware sampling strategy and successfully train a ResNet-driven Siamese tracker with significant performance gain. Moreover, we propose a new model architecture to perform layer-wise and depth-wise aggregations, which not only further improves the accuracy but also reduces the model size. We conduct extensive ablation studies to demonstrate the effectiveness of the proposed tracker, which obtains currently the best results on five large tracking benchmarks, including OTB2015, VOT2018, UAV123, LaSOT, and TrackingNet. Our model will be released to facilitate further researches.

\end{abstract}

\section{Introduction}
\label{sec_Introduction}
Visual object tracking has received increasing attention over the last decades and has remained a very active research direction. It has a large range of applications in diverse fields like visual surveillance \cite{ICPR10HumanTrack}, human-computer interactions \cite{ICIP12HandPose}, and augmented reality \cite{CVPR15TrackSLAM}. Although much progress has been made recently, it has still been commonly recognized as a very challenging task due to numerous factors such as illumination variation, occlusion, and background clutters, to name a few \cite{PAMI15OTB}.

Recently, the Siamese network based trackers \cite{CVPR16SINT, ECCV16SiamFC, ECCV16GOTURN, arXiv17DCFNet, CVPR17CFNet, CVPR18SiamRPN, CVPR18RASTrack, ECCV18DaSiamRPN, IJCAI18EDCF} have drawn much attention in the community. These Siamese trackers formulate the visual object tracking problem as learning a general similarity map by cross-correlation between the feature representations learned for the target template and the search region. To ensure tracking efficiency, the offline learned Siamese similarity function is often fixed during the running time \cite{CVPR16SINT, ECCV16SiamFC, ECCV16GOTURN}. The CFNet tracker \cite{CVPR17CFNet} and DSiam tracker \cite{ICCV17DSiam} update the tracking model via a running average template and a fast transformation module, respectively. The SiamRNN tracker \cite{CVPR18SiamRPN} introduces the region proposal network \cite{CVPR18SiamRPN} after the Siamese network and performs joint classification and regression for tracking. The DaSiamRPN tracker \cite{ECCV18DaSiamRPN} further introduces a distractor-aware module and improves the discrimination power of the model.

Although the above Siamese trackers have obtained outstanding tracking performance, especially for the well-balanced accuracy and speed, even the best performed Siamese trackers, such as SiamPRN, the accuracy still has a notable gap with the state-of-the-arts \cite{CVPR17ECO} on tracking benchmarks like OTB2015 \cite{PAMI15OTB}. We observe that all these trackers have built their network upon architecture similar to AlexNet \cite{NIP12AlexNet} and tried several times to train a Siamese tracker with more sophisticated architecture like ResNet \cite{CVPR16ResNet} yet with no performance gain. Inspired by this observation, we perform an analysis of existing Siamese trackers and find the core reason comes from the destroy of the strict translation invariance. Since the target may appear at any position in the search region, the learned feature representation for the target template should stay spatial invariant, and we further theoretically find that, among modern deep architectures, only the zero-padding variant of AlexNet satisfies this spatial invariance restriction.

To overcome this restriction and drive the Siamese tracker with more powerful deep architectures, through extensive experimental validations, we introduce a simple yet effective sampling strategy to break the spatial invariance restriction of the Siamese tracker. We successfully train a SiamRPN~\cite{CVPR18SiamRPN} based tracker using the ResNet as a backbone network and obtain significant performance improvements. Benefiting from the ResNet architecture, we propose a layer-wise feature aggravation structure for the cross-correlation operation, which helps the tracker to predict the similarity map from features learned at multiple levels. By analyzing the Siamese network structure for cross-correlations, we find that its two network branches are highly imbalanced in terms of parameter number; therefore we further propose a depth-wise separable correlation structure which not only greatly reduces the parameter number in the target template branch, but also stabilizes the training procedure of the whole model.
In addition, an interesting phenomena is observed that objects in the same categories have high response on the same channels while responses of the rest channels are suppressed. The orthogonal property may also improve the tracking performance.

To summarize, the main contributions of this work are listed below in fourfold:
\vspace{-3mm}
\begin{itemize}
\setlength{\itemsep}{5pt}
\setlength{\parsep}{5pt}
\setlength{\parskip}{0pt}

\item We provide a deep analysis of Siamese trackers and prove that when using deep networks the decrease in accuracy comes from the destroy of the strict translation invariance.
\item We present a simple yet effective sampling strategy to break the spatial invariance restriction which successfully trains Siamese tracker driven by a ResNet architecture.
\item We propose a layer wise feature aggregation structure for the cross-correlation operation, which helps the tracker to predict the similarity map from features learned at multiple levels.
\item We propose a depth-wise separable correlation structure to enhance the cross-correlation to produce multiple similarity maps associated with different semantic meanings.
\end{itemize}

Based on the above theoretical analysis and technical contributions, we have developed a highly effective and efficient visual tracking model which establishs a new state-of-the-art in terms of tracking accuracy, while running efficiently at 35 FPS. The proposed tracker, referred as \emph{SiamRPN++}, consistently obtains the best tracking results on five of the largest tracking benchmarks, including OTB2015 \cite{PAMI15OTB}, VOT2018 \cite{VOT18Results}, UAV123 \cite{ECCV16UAV123}, LaSOT \cite{ARX18LaSOT}, and TrackingNet \cite{ECCV18trackingnet}. Furthermore, we propose a fast variant of our tracker using MobileNet\cite{ARX17mobile} backbone that maintains competitive performance, while running at 70 FPS. To facilitate further studies on the visual tracking direction, we will release the source code and trained models of the SiamRPN++ tracker.

\section{Related Work}
\label{sec_RelatedWork}
In this section, we briefly introduce recent trackers, with a special focus on the Siamese network based trackers \cite{CVPR16SINT, ECCV16SiamFC}. Besides, we also describe the recent developments of deep architectures.

Visual tracking has witnessed a rapid boost in the last decade due to the construction of new benchmark datasets \cite{CVPR13OTB, PAMI15OTB, VOT16Results, VOT18Results, ARX18LaSOT, ECCV18trackingnet} and improved methodologies \cite{PAMI15KCF, ICCVW15RAJSSC, ICCV15SRDCF, ICCVW15DeepSRDCF, CVPR15Muster, CVPR16MDNet, ECCV16CCOT, CVPR17ECO, CVPR18RASTrack, ECCV18DaSiamRPN, ECCV18SACFN}. The standardized benchmarks \cite{CVPR13OTB, PAMI15OTB, ARX18LaSOT} provide fair testbeds for comparisons with different algorithms. The annually held tracking challenges \cite{VOT15Results, VOT16Results, VOT17Results, VOT18Results} are consistently pushing forward the tracking performance. With these advancements, many promising tracking algorithms have been proposed. The seminal work by Bolme \etal \cite{CVPR10MOSSE} introduces the Convolution Theorem from the signal processing field into visual tracking and transforms the object template matching problem into a correlation operation in the frequency domain. Own to this transformation, the correlation filter based trackers gain not only highly efficient running speed, but also increase accuracy if proper features are used \cite{PAMI15KCF, ICIP15JSSC, ICCVW15RAJSSC, CVPR14CN, ICCV15SRDCF}. With the wide adoption of deep learning models in visual tracking, tracking algorithms based on correlation filter with deep feature representations \cite{ECCV16CCOT, CVPR17ECO} have obtained the state-of-the-art accuracy in popular tracking benchmarks \cite{CVPR13OTB, PAMI15OTB} and challenge \cite{VOT15Results, VOT16Results, VOT17Results}.

Recently, the Siamese network based trackers have received significant attentions for their well-balanced tracking accuracy and efficiency \cite{CVPR16SINT,ECCV16SiamFC, ECCV16GOTURN, arXiv17DCFNet, CVPR17CFNet, ICCV17DSiamFC, CVPR18SiamRPN, CVPR18RASTrack, ECCV18DaSiamRPN, IJCAI18EDCF}. These trackers formulate visual tracking as a cross-correlation problem and are expected to better leverage the merits of deep networks from end-to-end learning. In order to produce a similarity map from cross-correlation of the two branches, they train a Y-shaped neural network that joins two network branches, one for the object template and the other for the search region. Additionally, these two branches can remain fixed during the tracking phase \cite{CVPR16SINT,ECCV16SiamFC, ECCV16GOTURN, CVPR18RASTrack, CVPR18SiamRPN, ECCV18DaSiamRPN} or updated online to adapt the appearance changes of the target \cite{arXiv17DCFNet, CVPR17CFNet, ICCV17DSiamFC}. The currently state-of-the-art Siamese trackers \cite{CVPR18SiamRPN, ECCV18DaSiamRPN} enhance the tracking performance by a region proposal network after the Siamese network and produce very promising results. However, on the OTB benchmark \cite{PAMI15OTB}, their tracking accuracy still leaves a relatively large gap with state-of-the-art deep trackers like ECO \cite{CVPR17ECO} and MDNet \cite{CVPR16MDNet}.

With the proposal of modern deep architecture AlexNet by Alex \etal \cite{NIP12AlexNet} in 2012, the studies of the network architectures are rapidly growing and many sophisticated deep architectures are proposed, such as VGGNet \cite{ICLR15VGG}, GoogleNet \cite{CVPR15GoogleNet}, ResNet \cite{CVPR16ResNet} and MobileNet \cite{ARX17mobile}. These deep architectures not only provide deeper understanding on the design of neural networks, but also push forwards the state-of-the-arts of many computer vision tasks like object detection \cite{CVPR18MegDet}, image segmentation \cite{ECCV18SegEDASC}, and human pose estimation \cite{ECCV18PoseDLC}. 
In deep visual trackers, the network architecture usually contains no more than five constitutional layers tailored from AlexNet or VGGNet. This phenomenon is explained that shallow features mostly contribute to the accurate localization of the object \cite{ARX17STrackerAnalysis}. 
In this work, we argue that the performance of Siamese trackers can significantly get boosted using deeper models if the model is properly trained with the whole Siamese network.

\section{Siamese Tracking with Very Deep Networks}
\label{sec_SiamResNet}
  
The most important finding of this work is that the performance of the Siamese network based tracking algorithm can be significantly boosted if it is armed with much deeper networks. However, simply training a Siamese tracker by directly using deeper networks like ResNet does not obtain the expected performance improvement. We find the underlying reason largely involves the intrinsic restrictions of the Siamese trackers, Therefore, before the introduction of the proposed SiamRPN++ model, we first give a deeper analysis on the Siamese networks for tracking.

\subsection{Analysis on Siamese Networks for Tracking}
\label{sec_analysis}
    
The Siamese network based tracking algorithms \cite{CVPR16SINT, ECCV16SiamFC} formulate visual tracking as a cross-correlation problem and learn a tracking similarity map from deep models with a Siamese network structure, one branch for learning the feature presentation of the target, and the other one for the search area. The target patch is usually given in the first frame of the sequence and can be viewed as an exemplar $\mathbf{z}$. The goal is to find the most similar patch (instance) from following frame $\mathbf{x}$ in a semantic embedding space $\phi(\cdot)$:
\begin{equation}
f(\mathbf{z}, \mathbf{x})=\phi(\mathbf{z})\ast \phi(\mathbf{x})+b,
\label{SiamProb}
\end{equation}
where $b$ is used to model the offset of the similarity value.

This simple matching function naturally implies two \emph{intrinsic} restrictions in designing a Siamese tracker.
\vspace{-1mm}
\begin{itemize}
\setlength{\itemsep}{-1pt}
\setlength{\parsep}{0pt}
\setlength{\parskip}{0pt}
\item The contracting part and the feature extractor used in Siamese trackers have an intrinsic restriction for \emph{strict translation invariance}, $f(\mathbf{z}, \mathbf{x}[\bigtriangleup\tau_j])=f(\mathbf{z}, \mathbf{x})[\bigtriangleup\tau_j]$, where $[\bigtriangleup\tau_j]$ is the translation shift sub window operator, which ensures the efficient training and inference.
\item The contracting part has an intrinsic restriction for \emph{structure symmetry}, \ie~$f(\mathbf{z}, \mathbf{x}')=f(\mathbf{x}', \mathbf{z})$, which is appropriate for the similarity learning.
\vspace{0mm}
\end{itemize}

After detailed analysis, we find the core reason for preventing Siamese tracker using deep network is related to these two aspects. Concretely speaking, one reason is that padding in deep networks will destroy the strict translation invariance. The other one is that RPN requires \emph{asymmetrical} features for classification and regression. We will introduce spatial aware sampling strategy to overcome the first problem, and discuss the second problem in Sect. \ref{sec:dw}.

Strict translation invariance only exists in no padding network such as modified AlexNet \cite{ECCV16SiamFC}. Previous Siamese based Networks \cite{ECCV16SiamFC, arXiv17DCFNet, CVPR17CFNet, CVPR18SiamRPN, ECCV18DaSiamRPN} are designed to be shallow to satisfy this restriction.
However, if the employed networks are replaced by modern networks like ResNet or MobileNet, padding is inevitable to make the network going deeper, which destroys the strict translation invariance restriction. Our hypothesis is that the violation of this restriction will lead to a spatial bias.
 
\begin{figure}[t]
\begin{center}
\includegraphics[width=0.99\columnwidth]{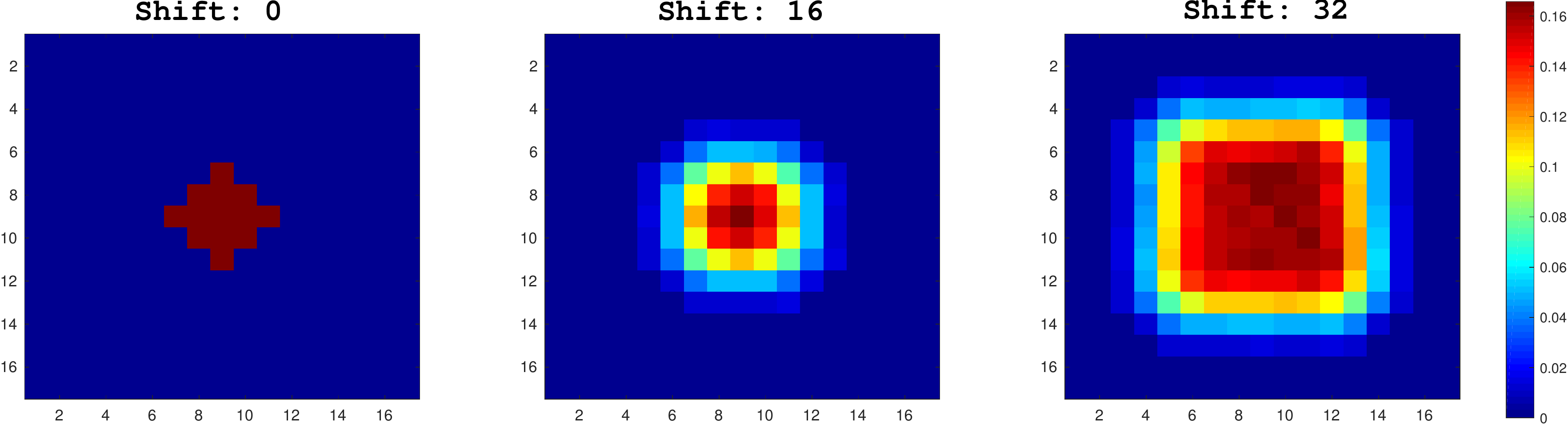}
\end{center}
\vspace{-3mm}
\caption{Visualization of prior probabili ties of positive samples when using different random translations. The distributions become more uniform after random translations within $\pm32$ pixels.}
\vspace{-3mm}
\label{fig:spbias2}
\end{figure}

We test our hypothesis by simulation experiments on a network with padding. 
Shift is defined as the max range of translation generated by a uniform distribution in data augmentation. 
Our simulation experiments are performed as follows. 
First, targets are placed in the center with different shift ranges (0, 16 and 32) in three sepreate training experiments.
After convergence, we aggregate the heatmaps generated on test dataset and then visualize the results in Fig. \ref{fig:spbias2}.
In the first simulation with zero shift, the probabilities on the border area are degraded to zero. It shows that a strong center bias is learned despite of the appearances of test targets.
The other two simulations show that increasing shift ranges will gradually prevent model collapse into this trivial solution.
The quantitative results illustrate that the aggregated heatmap of 32-shift is closer to the location distribution of test objects.
It proves that the spatial aware sampling strategy effectively alleviate the break of strict translation invariance property caused by the networks with padding.

\begin{figure}[t]
\begin{center}
\includegraphics[width=0.99\columnwidth]{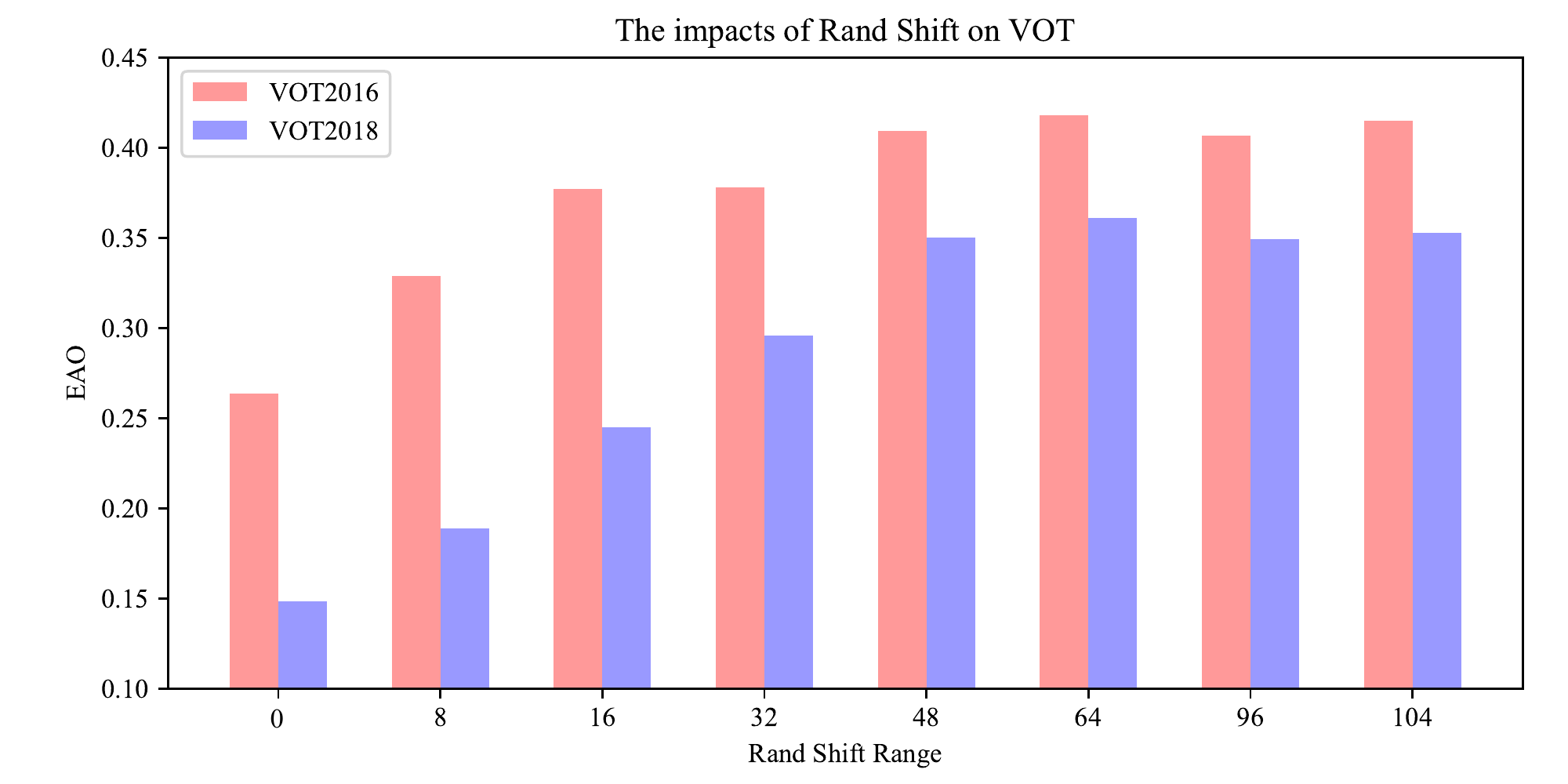}
\end{center}
\vspace{-4mm}
\caption{The impacts of the random translation on VOT dataset.}
\vspace{-5mm}
\label{fig:shifteao}
\end{figure}

 \begin{figure*}[t]
 \begin{center}
 \includegraphics[width=1.9\columnwidth]{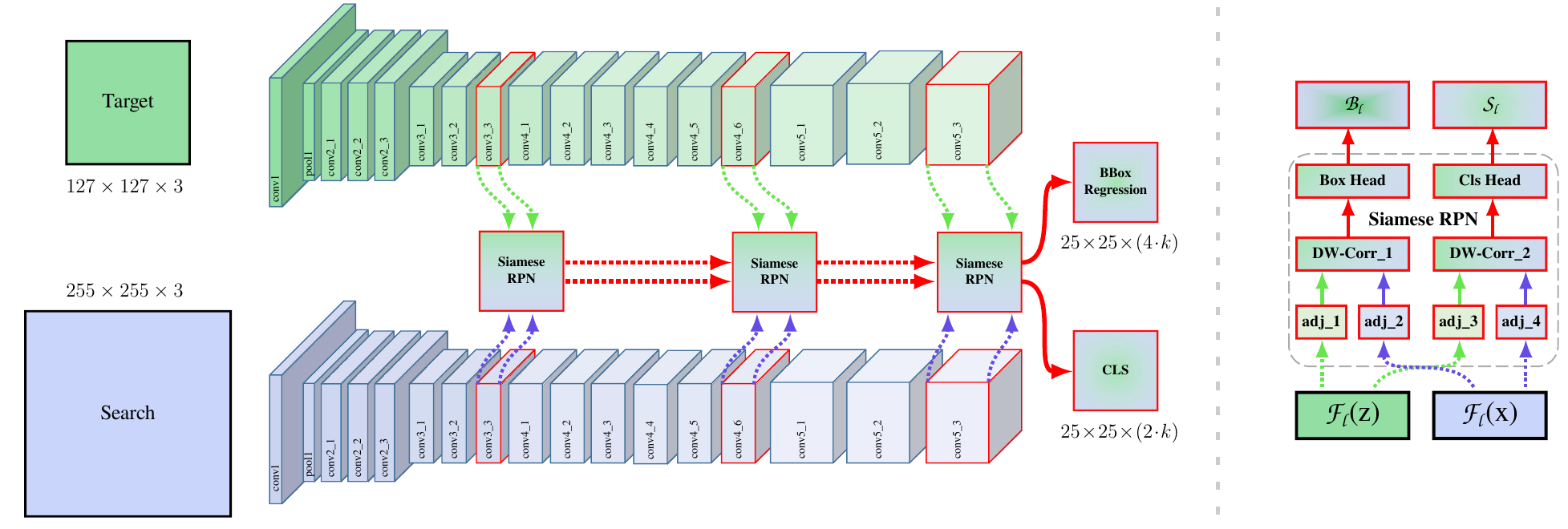}
 \end{center}
 \vspace{-3mm}
 \caption{Illustration of our proposed framework. Given a target template and search region, the network ouputs a dense prediction by fusion the outputs from multiple Siamese Region Proposal (SiamRPN) blocks. Each SiamRPN block is shown on right.}
 \vspace{-2mm}
 \label{fig:framework}
 \end{figure*}

To avoid putting a strong center bias on objects, we train SiamRPN with a ResNet-50 backbone by the spatial aware sampling strategy. 
As shown in Fig. \ref{fig:shifteao}, the performance with zero shift reduced to 0.14 on VOT2018, a suitable shift ($\pm 64$ pixels) is vital for training a deep Siamese tracker.

\subsection{ResNet-driven Siamese Tracking}
\label{sec:backbone}
Based on the above analyses, the influence of center bias can be eliminated. Once we eliminate the learning bias to the center location, any off-the-shelf networks~(\emph{e.g.}, MobileNet, ResNet) can be utilized to perform visual tracking after domain adaptation. Moreover, we can adaptively construct the network topology and unveil the performance of \emph{deep} network for visual tracking.

In this subsection, we will discuss how to transfer a deep network into our tracking algorithms. In particular, we conduct our experiments mainly focusing on ResNet-50 \cite{CVPR16ResNet}. 
The original ResNet has a large stride of 32 pixels, which is not suitable for dense Siamese network prediction. As shown in Fig.\ref{fig:framework}, we reduce the effective strides at the last two block from 16 pixels and 32 pixels to 8 pixels by modifying the $conv4$ and $conv5$ block to have unit spatial stride, and also increase its receptive field by dilated convolutions~\cite{CVPR15FCN}. An extra $1\times1$ convolution layer is appended to each of block outputs to reduce the channel to 256.

Since the paddings of all layers are kept, the spatial size of the template feature increases to 15, which imposes a heavy computational burden on the correlation module. Thus we crop the center $7\times 7$ regions \cite{CVPR17CFNet} as the template feature where each feature cell can still capture the entire target region.

Following~\cite{CVPR18SiamRPN}, we use a combination of cross correlation layers and fully convolutional layers to assemble a head module for calculating classification scores (denoted by $\mathcal{S}$) and bounding box regressor (denoted by $\mathcal{B}$). The Siamese RPN blocks are denoted by $\mathcal{P}$. 

Furthermore, we find that carefully fine-tuning ResNet will boost the performance. By setting learning rate of ResNet extractor with 10 times smaller than RPN parts, the feature representation can be more suitable for tracking tasks. 
Different from traditional Siamese approaches, the parameters of the deep network are jointly trained in an end-to-end fashion. To the best of our knowledge, we are the first to achieve an end-to-end learning on a deep Siamese Network ($>$ 20 layers) for visual tracking.

\vspace{-2mm}
\subsection{Layer-wise Aggregation}

After utilizing deep network like ResNet-50, aggregating different deep layers becomes possible. 
Intuitively, visual tracking requires rich representations that span levels from low to high, scales from small to large, and resolutions from fine to coarse. Even with the depth of features in a convolutional network, a layer in isolation is not enough: compounding and aggregating these representations improve inference of recognition and localization. 

In the previous works which only use shallow networks like AlexNet, multi-level features cannot provide very different representations. However, different layers in ResNet are much more meaningful considering that the receptive field varies a lot.
Features from earlier layers will mainly focus on low level information such as color, shape, are essential for localization, while lacking of semantic information; Features from latter layers have rich semantic information that can be beneficial during some challenge scenarios like motion blur, huge deformation. The use of this rich hierarchical information is hypothesized to help tracking.
 
In our network, multi-branch features are extracted to collaboratively infer the target localization. As for ResNet-50, we explore multi-level features extracted from the last three residual block for our layer-wise aggregation. We refer these outputs as $\mathcal{F}_3(\mathbf{z})$, $\mathcal{F}_4(\mathbf{z})$, and $\mathcal{F}_5(\mathbf{z})$, respectively. As shown in Fig.~\ref{fig:framework}, the outputs of $conv3$, $conv4$, $conv5$ are fed into three Siamese RPN module individually.

Since the output sizes of the three RPN modules have the same spatial resolution, weighted sum is adopted directly on the RPN output. A weighted-fusion layer combines all the outputs. 
\begin{equation}
\mathcal{S}_{all} = \sum_{l=3}^{5}\alpha_i*\mathcal{S}_{l},\quad \mathcal{B}_{all} = \sum_{l=3}^{5}\beta_i*\mathcal{B}_{l}.
\end{equation}
The combination weights are separated for classification and regression since their domains are different. The weight is end-to-end optimized offline together with the network.

In contrast to previous works, our approach does not explicitly combine convolutional features, but learn classifiers and regressions separately. Note that with the depth of the backbone network significantly increased, we can achieve substantial gains from the sufficient diversity of visual-semantic hierarchies. 

\begin{figure}[t]
\begin{center}
\subfigure[Cross Correlation Layer]{\includegraphics[height=1.7cm]{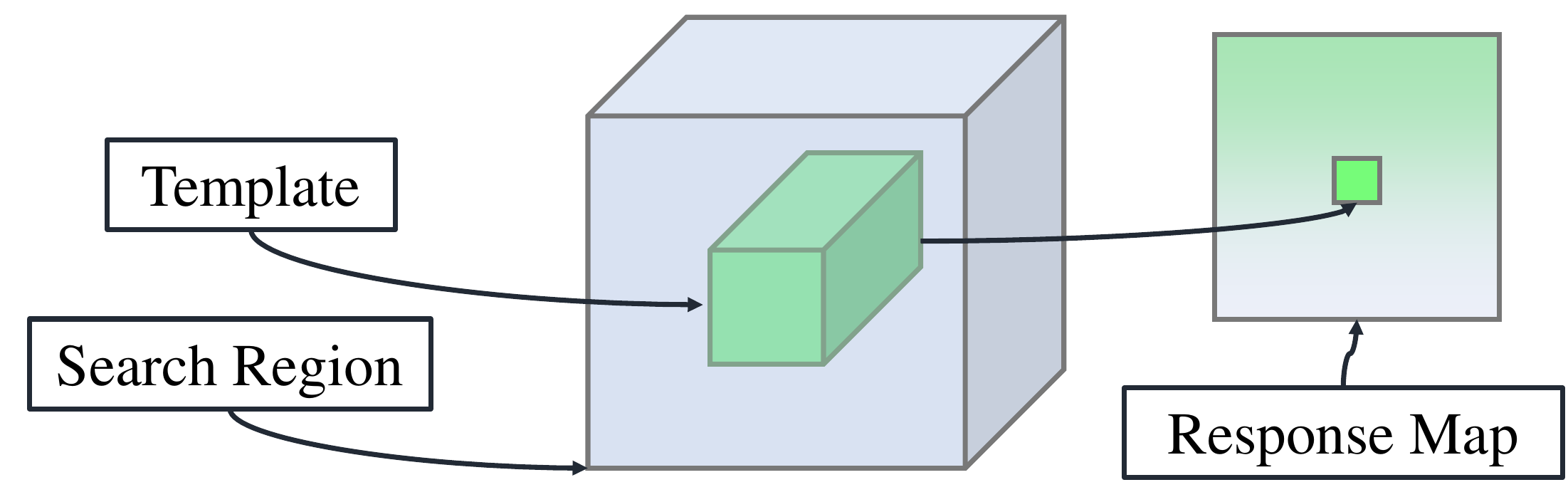}\label{cc}\vspace{-5mm}}
\subfigure[Up-Channel Cross Correlation Layer]{\includegraphics[height=2.2cm]{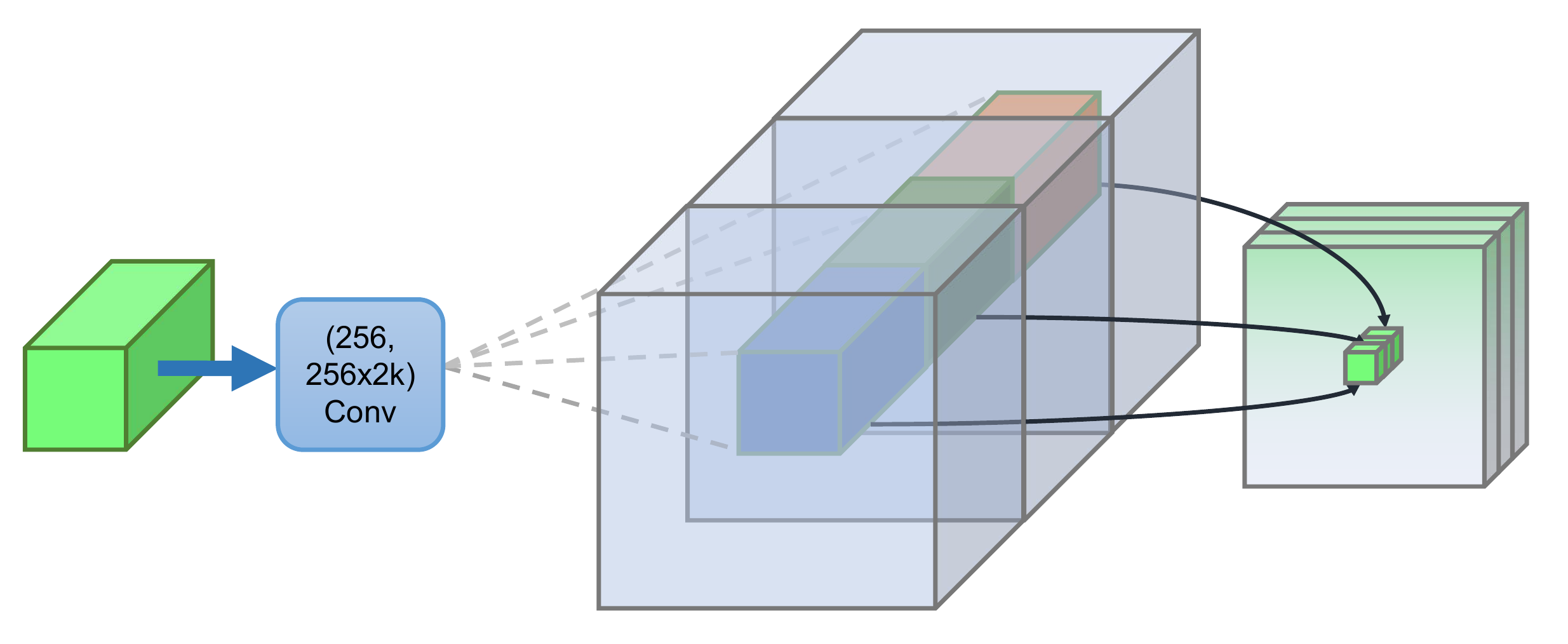}\label{up}}
\subfigure[Depth-wise Cross Correlation Layer]{\includegraphics[height=2cm]{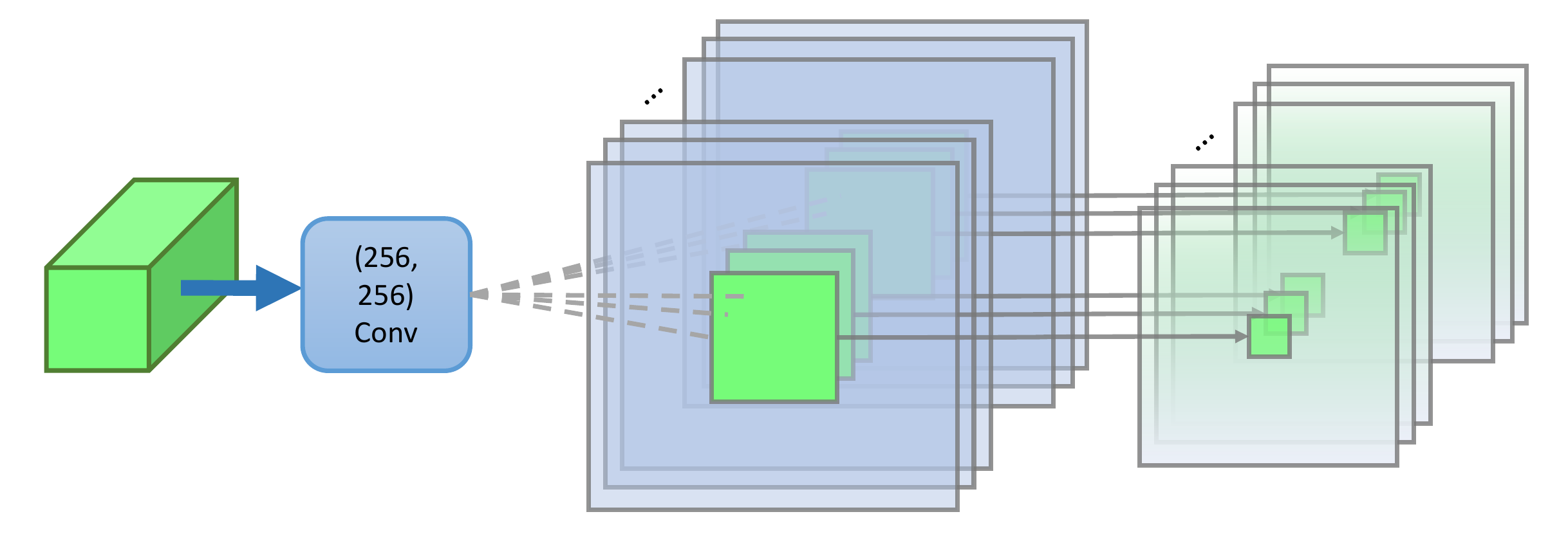}\label{dw}}
\end{center}
\vspace{-2mm}
\caption{Illustrations of different cross correlation layers. (a) Cross Correlation (XCorr) layer predicts a single channel similarity map between target template and search patches in SiamFC~\cite{ECCV16SiamFC}. (b) Up-Channel Cross Correlation (UP-XCorr) layer outputs a multi-channel correlation features by cascading a heavy convolutional layer with several independent XCorr layers in SiamRPN~\cite{CVPR18SiamRPN}. (c) Depth-wise Cross Correlation (DW-XCorr) layer predicts multi-channel correlation features between a template and search patches.}
\vspace{-1mm}
\label{fig:correaltion}
\end{figure}

\subsection{Depthwise Cross Correlation}
\label{sec:dw}
The cross correlation module is the core operation to embed two branches information. SiamFC~\cite{ECCV16SiamFC} utilizes a Cross-Correlation layer to obtain a single channel response map for target localization. In SiamRPN~\cite{CVPR18SiamRPN}, Cross-Correlation is extended to embed much higher level information such as anchors, by adding a huge convolutional layer to scale the channels (UP-Xcorr). The heavy up-channel module makes seriously imbalance of parameter distribution (\emph{i.e.} the RPN module contains 20M parameters while the feature extractor only contains 4M parameters in~\cite{CVPR18SiamRPN}), which makes the training optimization hard in SiamRPN.

In this subsection, we present a lightweight cross correlation layer, named Depthwise Cross Correlation (DW-XCorr), to achieve efficient information association. The DW-XCorr layer contains 10 times fewer parameters than the UP-XCorr used in SiamRPN while the performance is on par with it.

To achieve this, a conv-bn block is adopted to adjust features from each residual blocks to suit tracking task. Crucially, the bounding box prediction and anchor based classification both are \emph{asymmetrical}, which is different from SiamFC (See Sect. \ref{sec_analysis}). In order to encode the difference, the template branch and search branch pass two \emph{non-shared} convolutional layers. Then two feature maps with the same number of channels do the correlation operation channel by channel. Another conv-bn-relu block is appended to fuse different channel outputs. Finally, the last convolution layer for the output of classification or regression is appended.

By replacing cross-correlation to depthwise correlation, we can greatly reduce the computational cost and the memory usage. In this way, the numbers of parameters on the template and the search branches are balanced, resulting the training procedure more stable.

\begin{figure}[t]
\begin{center}
\includegraphics[width=0.99\linewidth]{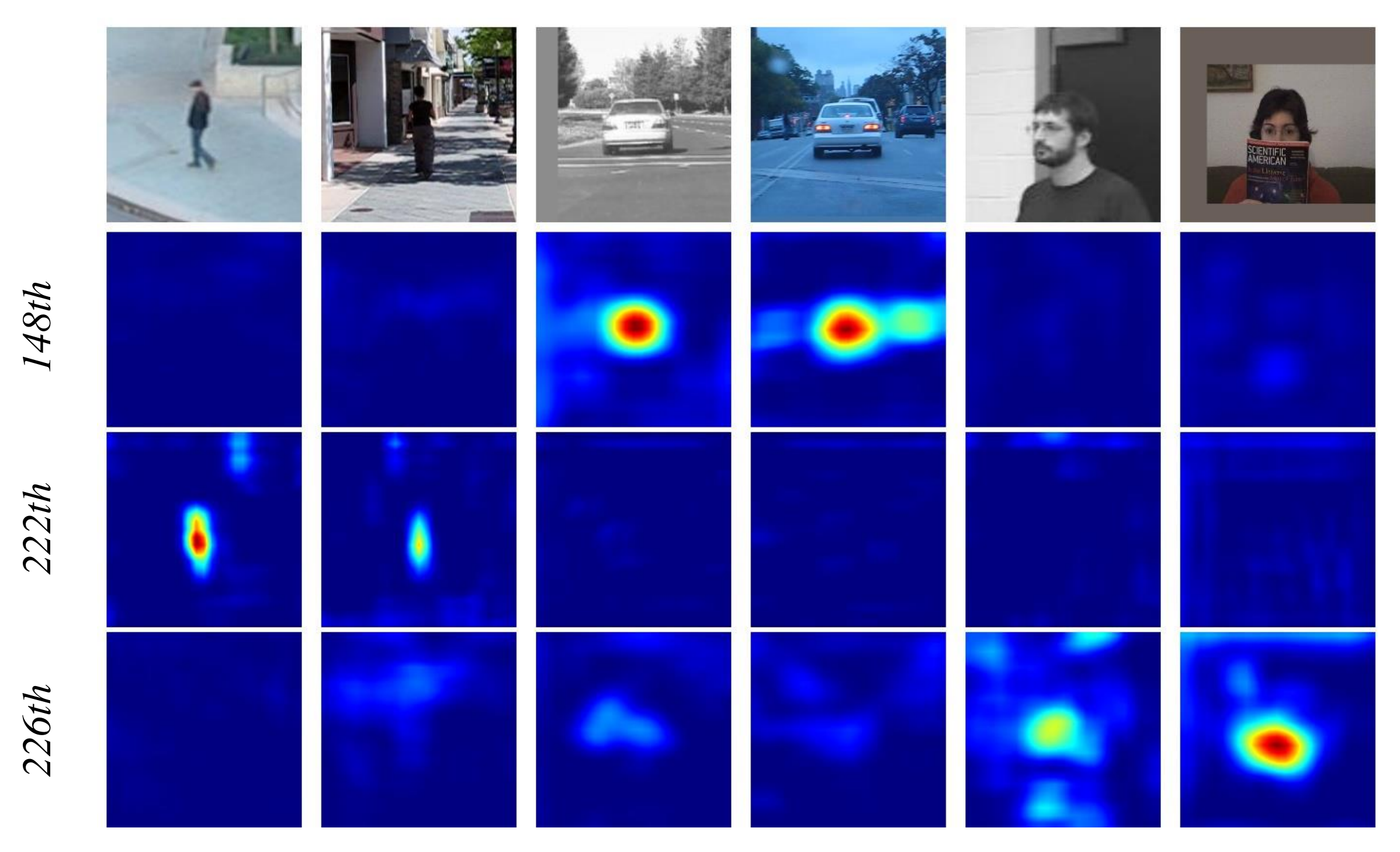}
\end{center}
\vspace{-1mm}
\caption{Channels of depthwise correlation output in \textit{conv4}. There are totally 256 channels in \textit{conv4}, however, only few of them have high response during tracking. Therefore we choose $148th$, $222th$, $226th$ channels as demonstration, which are $2nd$, $3rd$, $4th$ rows in the figure. The first row contains six corresponding search regions from OTB dataset \cite{PAMI15OTB}. Different channels represent different semantics, the $148th$ channel has high response on cars, while has low response on persons and faces. The $222th$ and $226th$ channel have high response on persons and faces, respectively.}
\vspace{-1mm}
 \label{fig:corr_res}
 \end{figure}
 
Furthermore, an interesting phenomena is illustrated in Fig.\ref{fig:corr_res}. The objects in the same category have high response on same channels (car in $148th$ channel, person in $222th$ channel, 
and face in $226th$ channel), while responses of the rest channels are suppressed. 
This property can be comprehended as the channel-wise features produced by the depthwise cross correlation are nearly orthogonal and each channel represents some semantic information. 
We also analyze the heatmaps when using the up-channel cross correlation and the reponse maps are less interpretable.

\section{Experimental Results}
\label{sec_experiment}

\subsection{Training Dataset and Evaluation}
\noindent
\textbf{Training}. The backbone network of our architecture~\cite{CVPR16ResNet} is pre-trained on ImageNet~\cite{IJCV15ImageNet} for image labeling, which has proven to be a very good initialization to other tasks~\cite{ICCV17MaskRCNN,CVPR15FCN}. We train the network on the training sets of COCO~\cite{ECCV2014COCO}, ImageNet DET~\cite{IJCV15ImageNet}, ImageNet VID, and YouTube-BoundingBoxes Dataset~\cite{CVPR17YTB} and to learn a generic notion of how to measure the similarities between general objects for visual tracking. In both training and testing, we use single scale images with 127 pixels for template patches and 255 pixels for searching regions.

\noindent
\textbf{Evaluation}. We focus on the short-term single object tracking on OTB2015~\cite{PAMI15OTB},  VOT2018~\cite{VOT18Results} and UAV123~\cite{ECCV16UAV123}. We use VOT2018-LT~\cite{VOT18Results} to evaulate the long-term setting. In the long-term tracking, the object may leave the field of view or become fully occluded for a long period, which are more challenging than short-term tracking. We also analyze the generalization of our method on LaSOT~\cite{ARX18LaSOT} and TrackingNet~\cite{ECCV18trackingnet}, two of the recent largest benchmarks for single object tracking.

\subsection{Implementation Details}
\noindent
\textbf{Network Architecture}. In experiments, we follow~\cite{ECCV18DaSiamRPN} for the training and inference settings. We attach two sibling convolutional layers to the stride-reduced ResNet-50 (Sect.~\ref{sec:backbone}) to perform proposal classification and bounding box regression with 5 anchors. Three randomly initialized $1\times1$ convolutional layers are attached to \textit{conv3}, \textit{conv4}, \textit{conv5} for reducing the feature dimension to 256.

\noindent
\textbf{Optimization}. SiamRPN++ is trained with stochastic gradient descent (SGD). We use synchronized SGD over 8 GPUs with a total of 128 pairs per minibatch (16 pairs per GPU), which takes 12 hours to converge. We use a warmup learning rate of 0.001 for first 5 epoches to train the RPN braches. For the last 15 epoches, the whole network is end-to-end trained with learning rate exponentially decayed from 0.005 to 0.0005. Weight decay of 0.0005 and momentum of 0.9 are used. The training loss is the sum of classification loss and the standard smooth $L_1$ loss for regression.

\subsection{Ablation Experiments}

\begin{figure}[t]
\begin{center}
\includegraphics[trim={0cm 0cm 0cm 0cm}, clip, width=0.99\columnwidth]{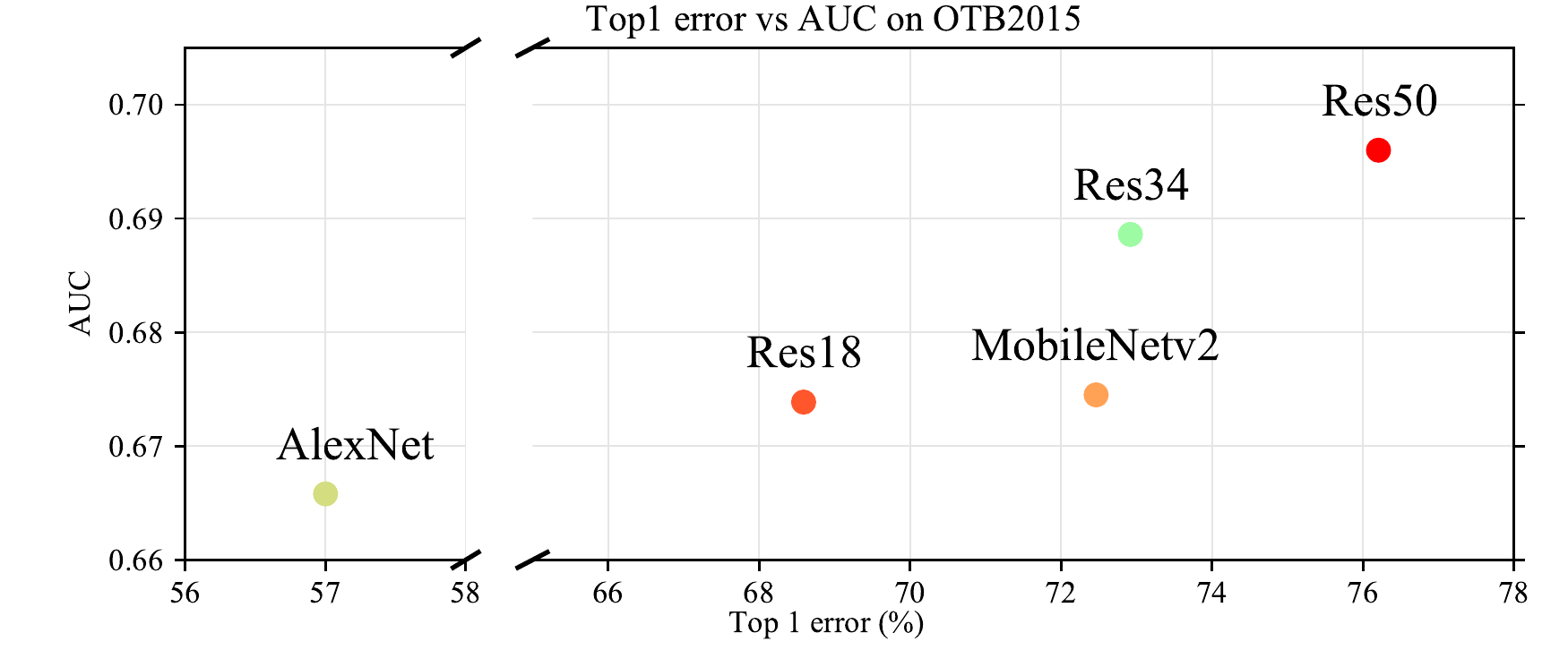}
\end{center}
\vspace{-4mm}
\caption{The Top-1 accuracy on ImageNet \textit{vs.} Expected Average Overlap (EAO) scores on OTB2015.}
\vspace{-2mm}
\label{fig:tpo1auc}
\end{figure}

\noindent
\textbf{Backbone Architecture}. The choice of feature extractor is crucial as the number of parameters and types of layers directly affect memory, speed, and performance of the tracker. We compare different network architectures for the visual tracking. Fig.~\ref{fig:tpo1auc} shows the performance of using AlexNet, ResNet-18, ResNet-34, ResNet-50, and MobileNet-v2 as backbones. We report performance by Area Under Curve (AUC) of success plot on OTB2015 with respect to the top1 accuracy on ImageNet. We observe that our SiamRPN++ can benefit from \emph{deeper} ConvNets.

Table \ref{tab:ablation} also illustrates that by replacing AlexNet to ResNet-50, the performance improves a lot on VOT2018 dataset. Besides, our experiments shows that finetuning the backbone part is critical, which yields a great improvement on tracking performance.

\renewcommand\arraystretch{1.0}
\setlength{\tabcolsep}{2.2pt}
\begin{table}[t]
\centering
\small
\begin{tabular}{c|ccc|c|c|c|c}
BackBone & L3 & L4 & L5 & Finetune & Corr & VOT2018 & OTB2015 \\
\shline
\multirow{2}{*}{AlexNet} & & & & & UP & 0.332 & 0.658\\
 & & & & & DW &  0.355 & 0.666 \\
\hline %
\multirow{2}{*}{ResNet-50} & \cmark & \cmark & \cmark &  & UP & 0.371 & 0.664 \\
 & \cmark & \cmark & \cmark & \cmark & UP & 0.390 & 0.684\\
\hline %
\multirow{6}{*}{ResNet-50} & \cmark & & & \cmark & DW  & 0.331 & 0.669 \\
 & & \cmark & & \cmark & DW  & 0.374 & 0.678 \\
 & & & \cmark & \cmark & DW  & 0.320 & 0.646 \\
 & \cmark & \cmark & & \cmark & DW  & 0.346 & 0.677 \\
 & \cmark & & \cmark & \cmark & DW  & 0.336 & 0.674 \\
 & & \cmark & \cmark & \cmark & DW  & 0.383 & 0.683 \\
\hline %
\multirow{2}{*}{ResNet-50} & \cmark & \cmark & \cmark &  & DW & 0.395 & 0.673 \\
 & \cmark & \cmark & \cmark & \cmark & DW & \first{0.414} & \first{0.696} \\
\end{tabular}
\caption{Ablation study of the proposed tracker on VOT2018 and OTB2015. L3, L4, L5 represent \textit{conv3},\textit{conv4},\textit{conv5}, respectively. Finetune represents whether the backbone is trained offline. Up/DW means Up channel correlation and depthwise correlation.}
\vspace{-3mm}
\label{tab:ablation}
\end{table}

\setlength{\tabcolsep}{.2em}
\begin{table*}[t] \small
\centering
\begin{tabular}{l| c c c c c c c c c c c c}
 & DLSTpp & DaSiamRPN & SA\_Siam\_R & CPT & DeepSTRCF & DRT & RCO & UPDT & SiamRPN & MFT & LADCF & \textbf{Ours}\\ %
\shline %
EAO $\uparrow$ & 0.325 & 0.326 & 0.337 & 0.339 & 0.345 & 0.356 & 0.376 & 0.378 & 0.383 & 0.385 & 0.389 & \color{red}\textbf{0.414}\\ %
Accuracy $\uparrow$ & 0.543 & 0.569 &  0.566 & 0.506 & 0.523 & 0.519 & 0.507 & 0.536 & 0.586 & 0.505 & 0.503 & \color{red}\textbf{0.600}\\ %
Robustness $\downarrow$ & 0.224 & 0.337 & 0.258 & 0.239 & 0.215 & 0.201 & 0.155 & 0.184 & 0.276 & \color{red}\textbf{0.140}    & 0.159 & 0.234\\ %
\hline %
AO $\uparrow$ & 0.495 & 0.398 & 0.429 & 0.379 & 0.436 & 0.426 & 0.384 & 0.454 &0.472 & 0.393 & 0.421 & \color{red}\textbf{0.498}\\
\end{tabular}
\caption{Comparison with the state-of-the-art in terms of expected average overlap (EAO), robustness (failure rate), and accuracy on the VOT2018 benchmark. We compare with the top-10 trackers and our baseline DaSiamRPN in the competition. Our tracker obtains a significant relative gain of $6.4\%$ in EAO, compared to the top-ranked method (LADCF).} 
\vspace{-5mm}
\label{tab:vot18}
\end{table*}

\noindent
\textbf{Layer-wise Feature Aggregation}.
To investigate the impact of layer-wise feature aggregation, first we train three variants with single RPN on ResNet-50. We empirically found that $conv4$ alone can achieve a competitive performance with $0.374$ in EAO, while deeper layer and shallower layer perform with $4\%$ drops. 
Through combining two branches, $conv4$ and $conv5$ gains improvement, however no improvement is observed on the other two combinations. Even though, the robustness has increased $10\%$, which is the key vulnerability of our tracker. It means that our tracker still has room for improvement. 
After aggregating all three layers, both accuracy and robustness steadily improve, with gains between $3.1\%$ and $1.3\%$ for VOT and OTB.
In total, layer-wise feature aggregation yields a $0.414$ EAO score on VOT2018, which is $4.0\%$ higher than that of the single layer baseline. 

\noindent
\textbf{Depthwise Correlation}.
We compare the original Up-Channel Cross Correlation layer with the proposed Depthwise Cross Correlation layer. As shown in the Table \ref{tab:ablation}, the proposed depthwise correlation gains $2.3\%$ improvement on VOT2018 and $0.8\%$ improvement on OTB2015, which demonstrates the importance of depthwise correlation. 
This is partly beacause a balanced parameter distribution of the two branches makes the learning process more stable, and converges better.

\subsection{Comparison with the state-of-the-art}

\begin{figure}
\begin{center}
\subfigure[Success Plot]{\includegraphics[trim={1cm 0cm 1.5cm 0cm}, clip, width=0.495\linewidth]{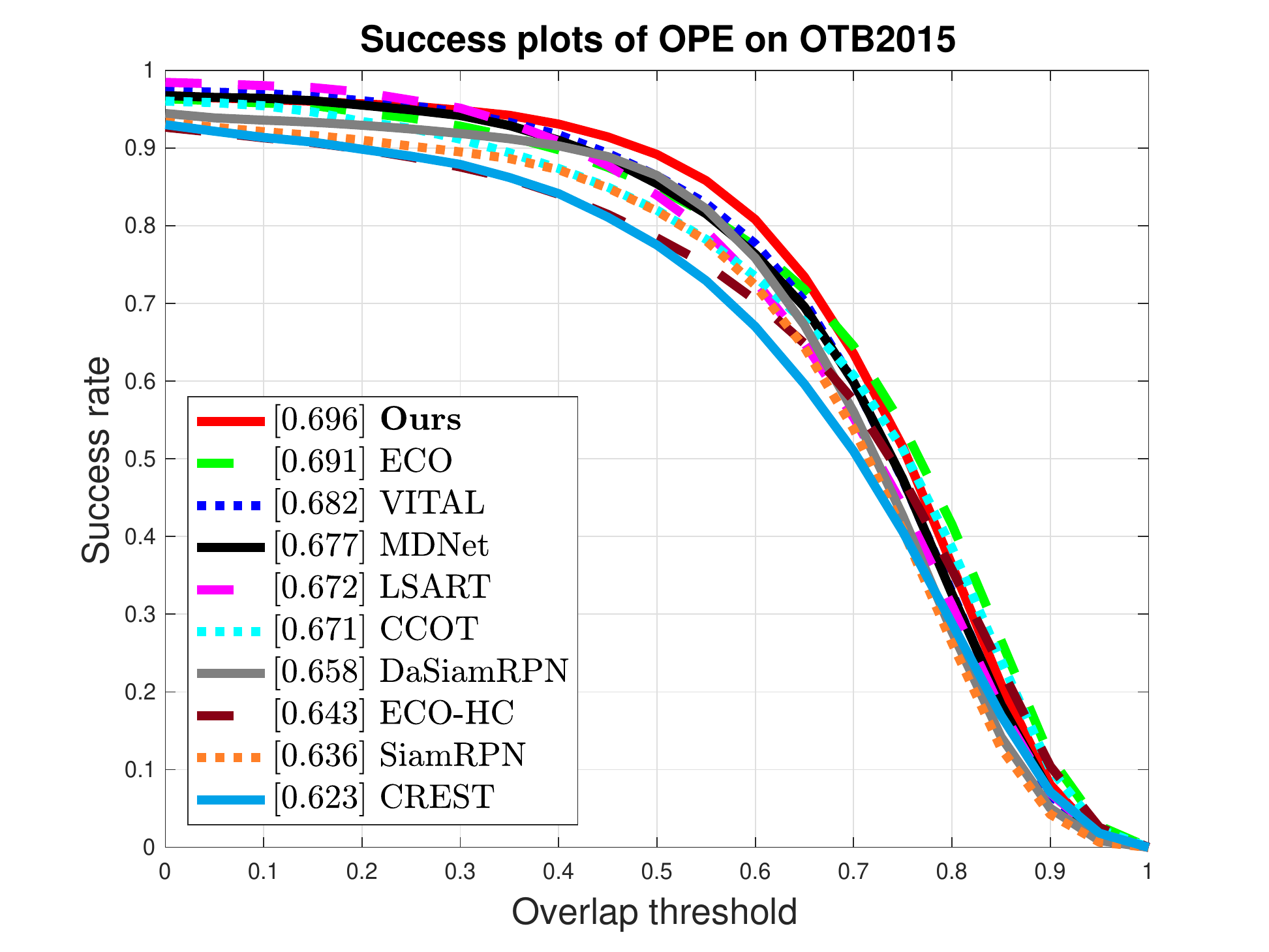}\label{Success Plot}}
\subfigure[Precision Plot]{\includegraphics[trim={1cm 0cm 1.5cm 0cm}, clip,width=0.495\linewidth]{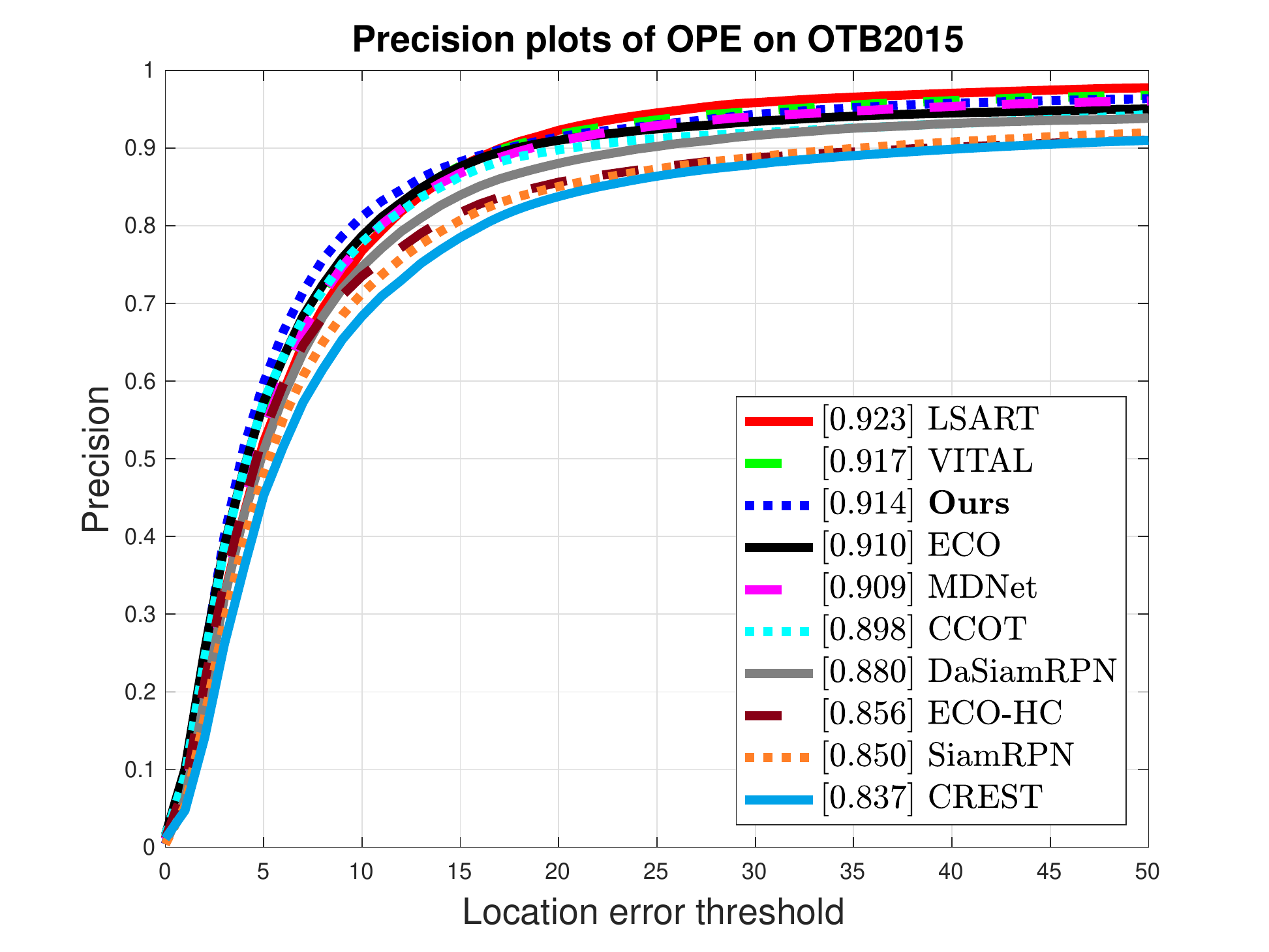}\label{Precision Plot}}
\end{center}
\vspace{-3mm}
\caption{Success and precision plots show a comparison of our tracker with state-of-the-art trackers on the OTB2015 dataset.}
\vspace{-3mm}
\label{fig:OTB}
\end{figure}

\paragraph{OTB-2015 Dataset.}
The standardized OTB benchmark \cite{PAMI15OTB} provides a fair testbed on robustness.
The Siamese based tracker formulate the tracking as one-shot detection task without any online update, thus resulting in inferior performance on this no-reset setting benchmark. 
However, we identify the limited representation from the \emph{shallow} network as the primary obstacle preventing Siamese based tracker from surpassing top-performing methods, such as C-COT variants~\cite{ECCV16CCOT, CVPR17ECO}.

We compare our SiamRPN++ tracker on the OTB2015 with the state-of-the-art trackers. Fig.~\ref{fig:OTB} shows that our SiamRPN++ tracker produces leading result in overlap success. Compared with the recent DaSiamRPN~\cite{ECCV18DaSiamRPN}, our SiamRPN++ improves $3.8\%$ in overlap and $3.4\%$ in precision from the considerably increased depth. Representations extracted from deep ConvNets are less sensitive to illumination and background clutter. And to the best of our knowledge, this is the first time that Siamese tracker can obtain the comparable performance with the state-of-the-art tracker on OTB2015 dataset.

\vspace{-5mm}
\paragraph{VOT2018 Dataset.} We test our SiamRPN++ tracker on the lastest VOT-2018 dataset~\cite{VOT18Results} in comparison with 10 state-of-the-art methods. The VOT-2018 public dataset is one of the most recent datasets for evaluating online model-free single object trackers, and includes 60 public sequences with different challenging factors. Following the evaluation protocol of VOT-2018, we adopt the Expected Average Overlap (EAO), Accuracy(A) and Robustness(R) and no-reset-based Average Overlap(AO) to compare different trackers. The detailed comparisons are reported in Table~\ref{tab:vot18}. 

From Table~\ref{tab:vot18}, we observe that the proposed SiamRPN++ method achieves the top-ranked performance on EAO, A and AO criteria. Especially, our SiamRPN++ tracker outperforms all existing trackers, including the VOT2018 challenge winner. Compared with the best tracker in the VOT2018 challenge (LADCF~\cite{VOT18Results}), the proposed method achieves a performance gain of $2.5\%$. In addition, our tracker achieves a substantial improvement over the challenge winner (MFT~\cite{VOT18Results}), with a gain of $9.5\%$ in accuracy.

In comparison with the baseline tracker DaSiamRPN, our approach yields substantial gains of $10.3\%$ on robustness, which is the common vulnerability of the Siamese Network based tracker against correlation filters method. Even though, due to the lack of adaption to the template, the robustness still has a gap with the state-of-art correlation filters methods~\cite{ECCV18UPDT} which relies on the online updating.
 
The One Pass Evaluation (OPE) is also adopted to evaluate trackers and the AO values are reported to demonstrate their performance. From the last row in Table~\ref{tab:vot18}, we can observe that our method achieves comparable performance compared to the DLSTpp~\cite{VOT18Results} and improves the DaSiamRPN~\cite{ECCV18DaSiamRPN} method by an absolute gain of $10.0\%$.

\begin{figure}[t]
\begin{center}
\includegraphics[trim={2cm 10.5cm 2cm 11cm}, clip, width=0.99\columnwidth]{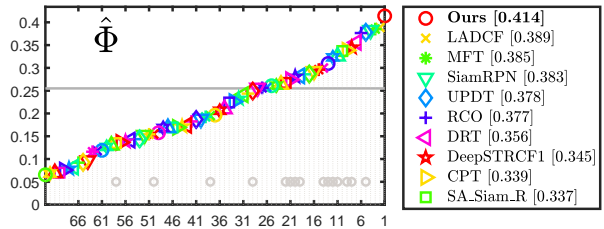}
\end{center}
\vspace{-3mm}
\caption{Expected averaged overlap performance on VOT2018.}
\vspace{-3mm}
\label{fig:votatrr}
\end{figure}

\begin{figure}
\begin{center}
\includegraphics[trim={0cm 0.5cm 0cm 0cm}, clip, width=0.99\columnwidth]{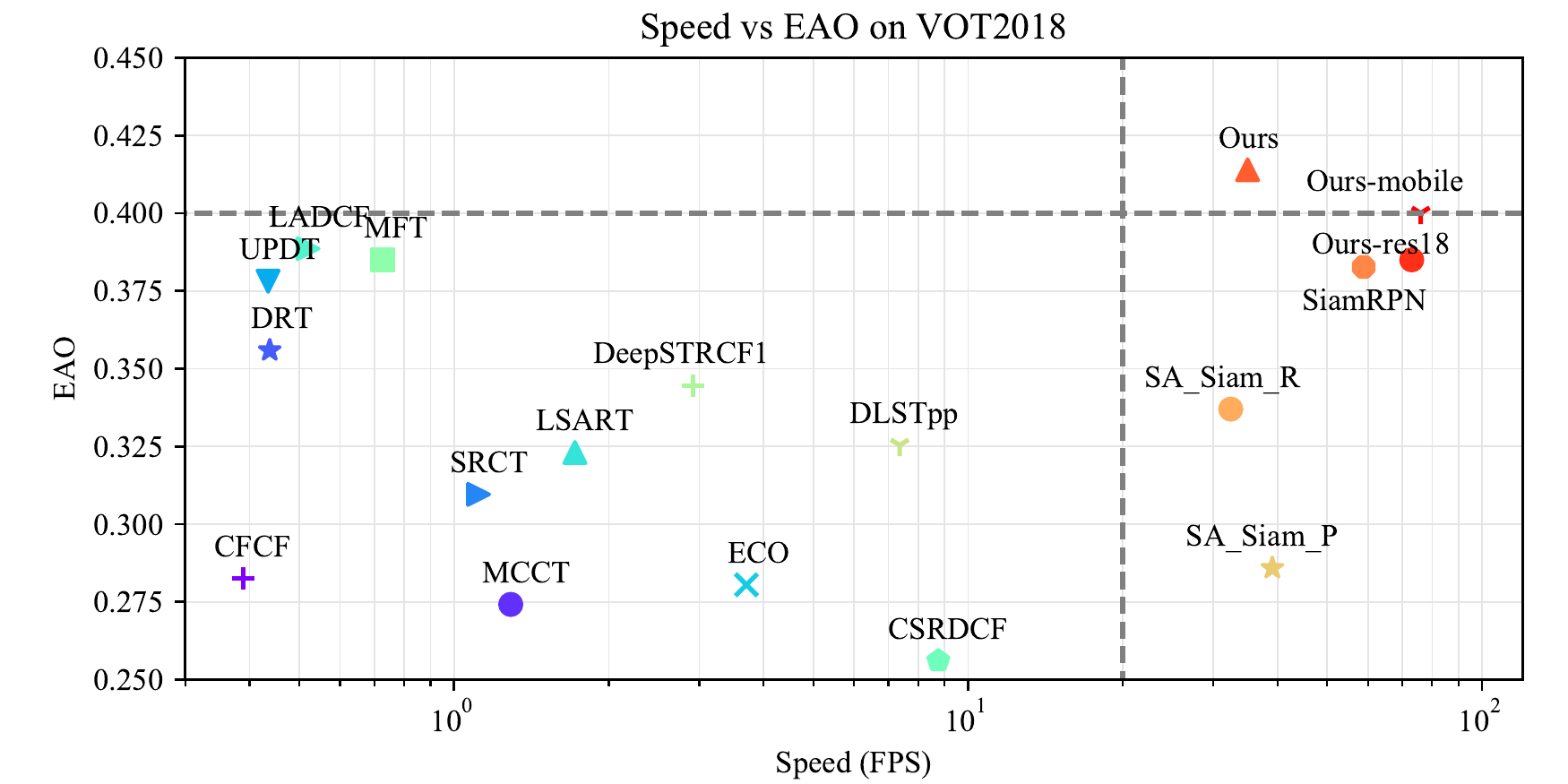}
\end{center}
\vspace{-3mm}
\caption{A comparison of the quality and the speed of state-of-the-art tracking methods on VOT2018. We visualize the Expected Average Overlap (EAO) with respect to the Frames-Per-Seconds (FPS). Note that the FPS axis is in the log scale. Two of our variants, which replace ResNet-50 backbone with ResNet-18~(Ours-res18) and MobileNetv2~(Ours-mobile), respectively.}
\vspace{-3mm}
\label{fig:speed}
\end{figure}

\begin{figure}
\begin{center}
\subfigure{\includegraphics[width=0.49\linewidth,height=0.49\linewidth]{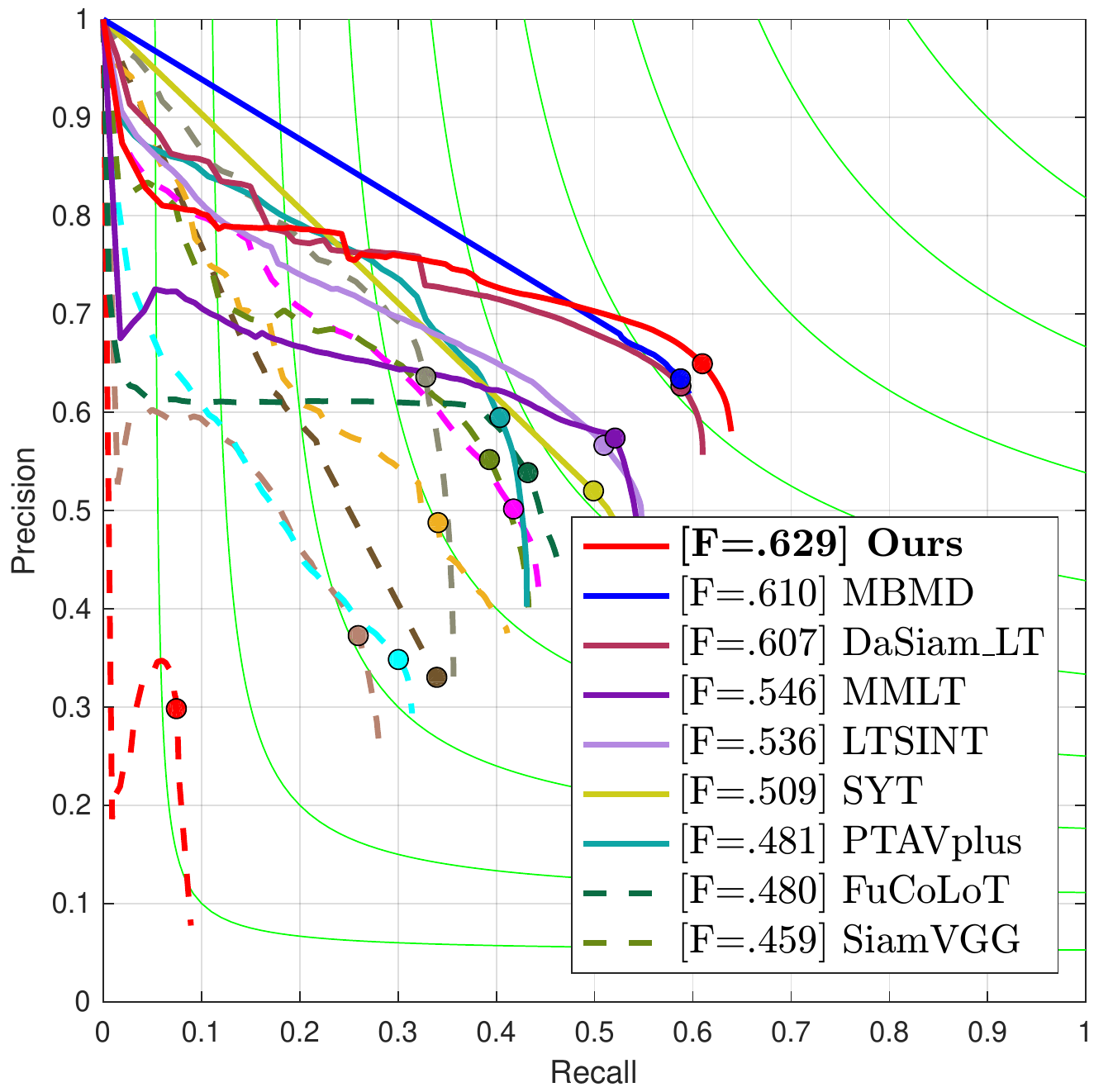}\label{fig:pr}}
\subfigure{\includegraphics[width=0.49\linewidth,height=0.49\linewidth]{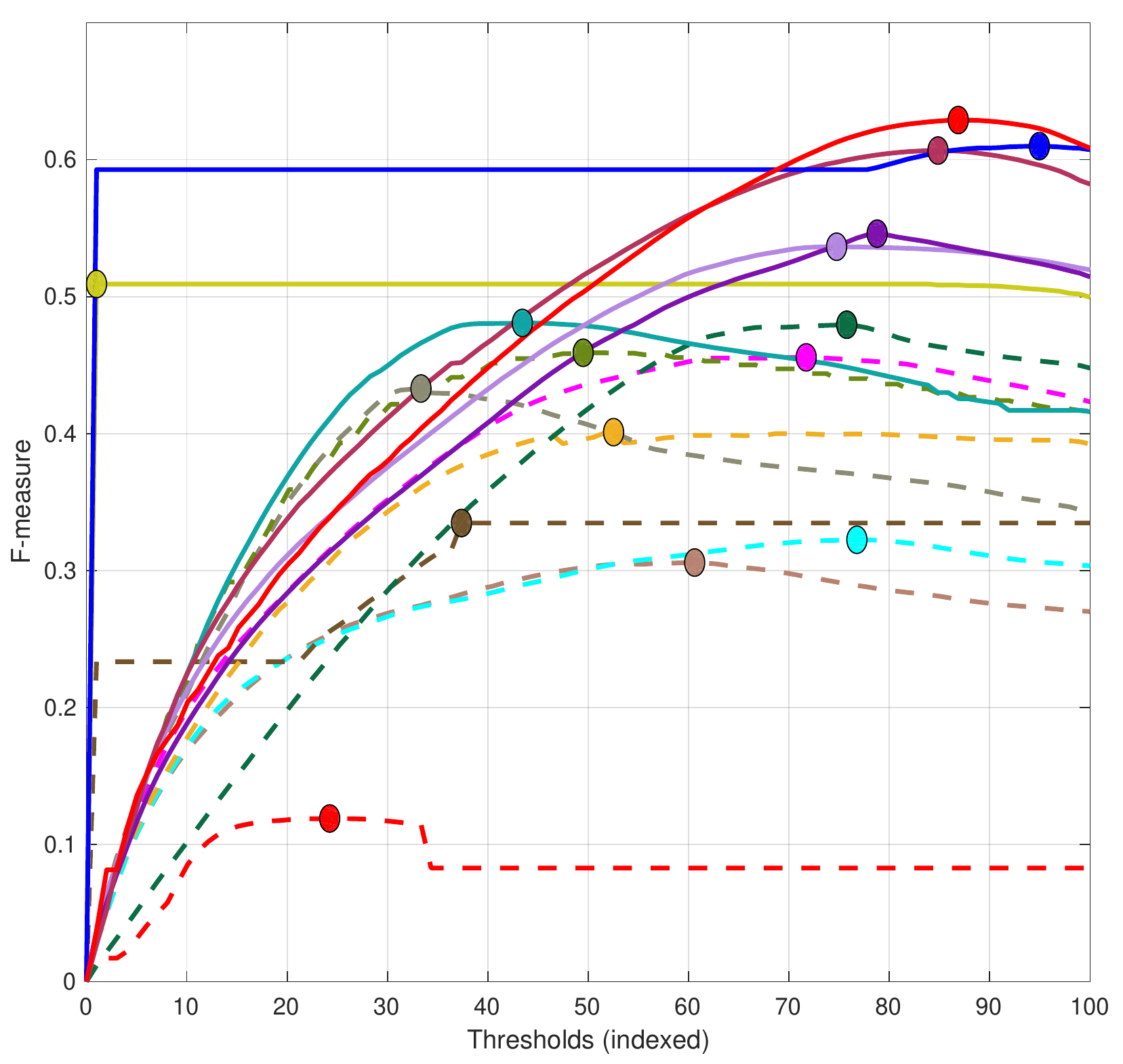}\label{fig:thr}}
\end{center}
\vspace{-3mm}
\caption{Long-term tracking performance. The average tracking precision-recall curves (left), the corresponding F-score curves (right). Tracker labels are sorted according to the F-score.}
\vspace{-3mm}
\label{fig:vot18lt}
\end{figure}

\noindent
\textbf{Accuracy vs. Speed}. 
In Fig.~\ref{fig:speed}, we visualize the EAO on VOT2018 with respect to the Frames-Per-Second (FPS). The reported speed is evaluated on a machine with an NVIDIA Titan Xp GPU, other results are provided by the VOT2018 official results.
From the plot, our SiamRPN++ achieves best performance, while still running at realtime speed(35 FPS). It is worth noting that two of our variants achieve nearly the same accuracy as SiamRPN++, while running at more than 70 FPS, which makes these two variants highly competitive.

\vspace{-2mm}
\paragraph{VOT2018 Long-term Dataset.}
In the latest VOT2018 challenge, a long-term experiment are newly introduced. It is composed of 35 long sequences, where targets may leave the field of view or become fully occluded for a long period. The performance measures are precision, recall and a combined F-score. We report all these metrics compared with the state-of-the-art trackers on VOT2018-LT.

As shown in the Fig. ~\ref{fig:vot18lt}, after equipping our tracker with the long term strategy, SiamRPN++ obtains $2.2\%$ gain from DaSiam\_LT, and outperforms the best tracker by $1.9\%$ in F-score. The powerful feature extracted by ResNet improves both TP and TR by $2\%$ absolutely from our baseline DaSiamRPN. Meanwhile, the long term version of SiamRPN++ is still able to run at 21 FPS, which is nearly 8 times faster than MBMD~\cite{VOT18Results}, the winner of VOT2018-LT.

\begin{figure}[t]
\begin{center}
\subfigure{\includegraphics[trim={1cm 0cm 1cm 0cm}, clip, width=0.495\linewidth]{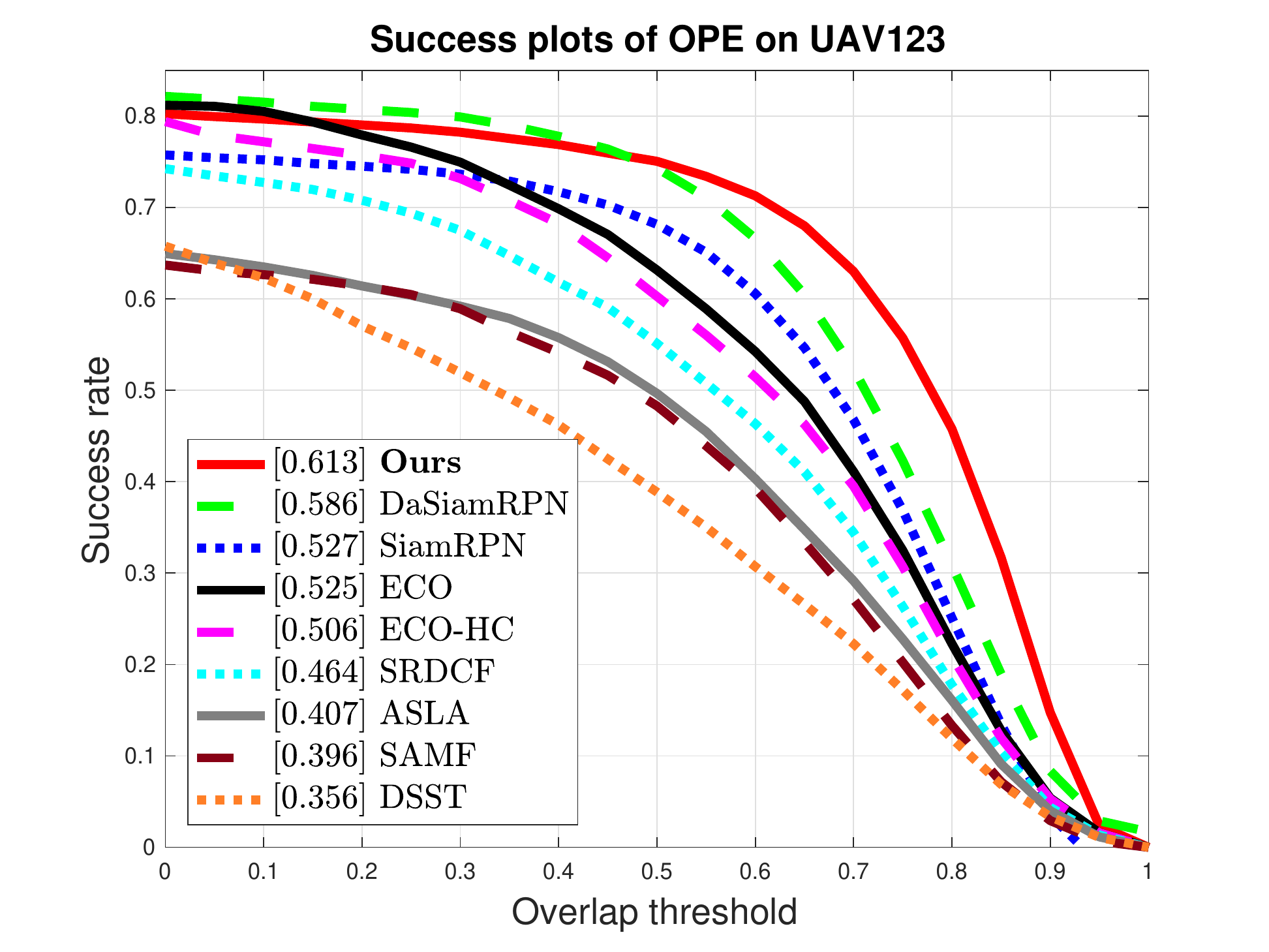}}
\subfigure{\includegraphics[trim={1cm 0cm 1cm 0cm}, clip, width=0.495\linewidth]{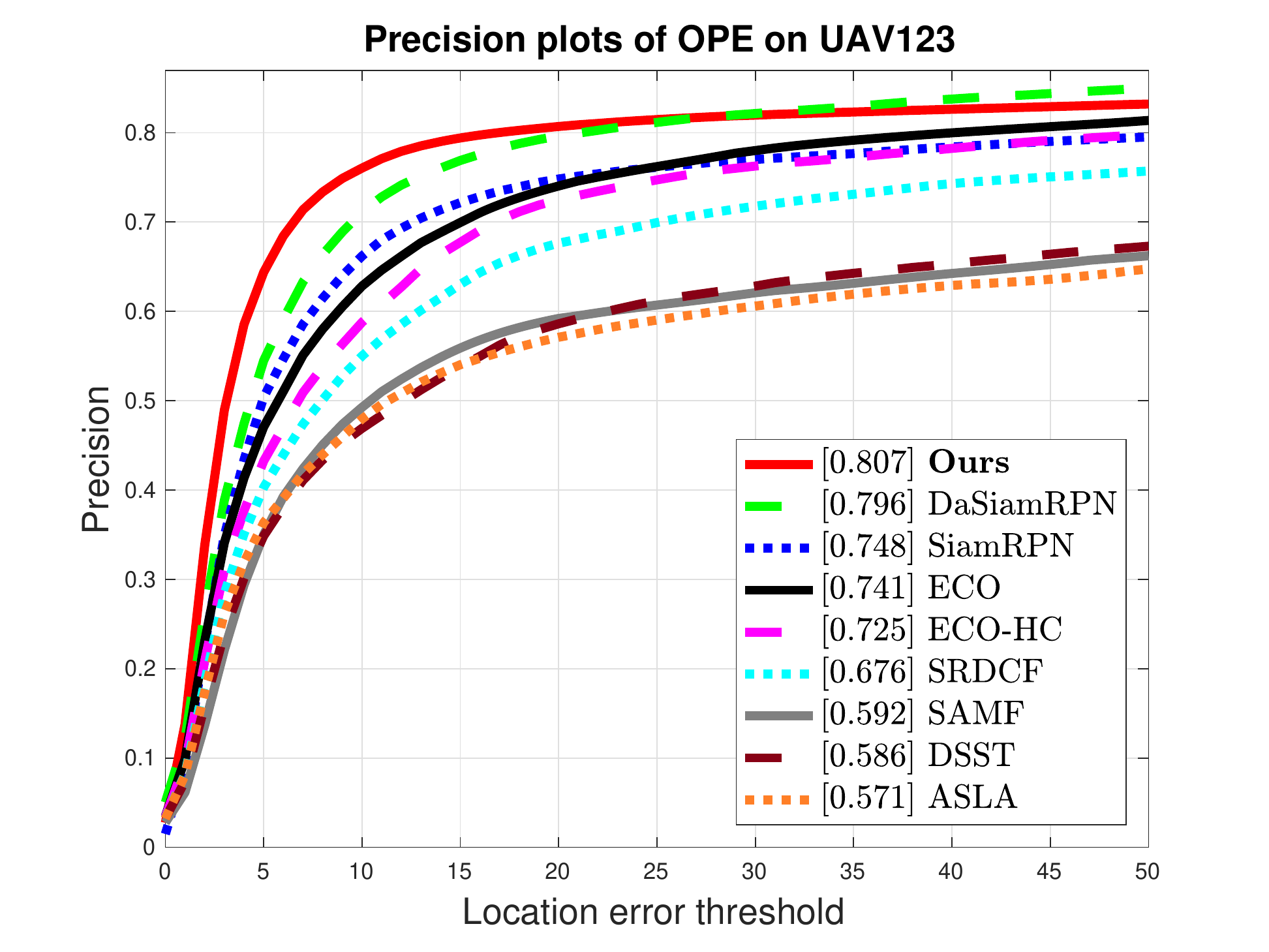}}
\end{center}
\vspace{-5mm}
\caption{Evaluation results of trackers on UAV123.}
\vspace{-6mm}
\label{fig:uav}
\end{figure}

\vspace{-5mm}
\paragraph{UAV123 Dataset.}
UAV123 dataset includes 123 sequences with average sequence length of 915 frames. Besides the recent trackers in \cite{ECCV16UAV}, ECO \cite{CVPR17ECO}, ECO-HC \cite{CVPR17ECO}, DaSiamRPN \cite{ECCV18DaSiamRPN}, SiamRPN \cite{CVPR18SiamRPN} are added on comparison. Fig.~\ref{fig:uav} illustrates the precision and success plots of the compared trackers. Specifically, our tracker achieves a success score of 0.613, which outperforms DaSiamRPN (0.586) and ECO (0.525) with a large margin.

\begin{figure}
\begin{center}
\subfigure{\includegraphics[trim={1cm 0cm 1cm 0cm}, clip, width=0.495\linewidth]{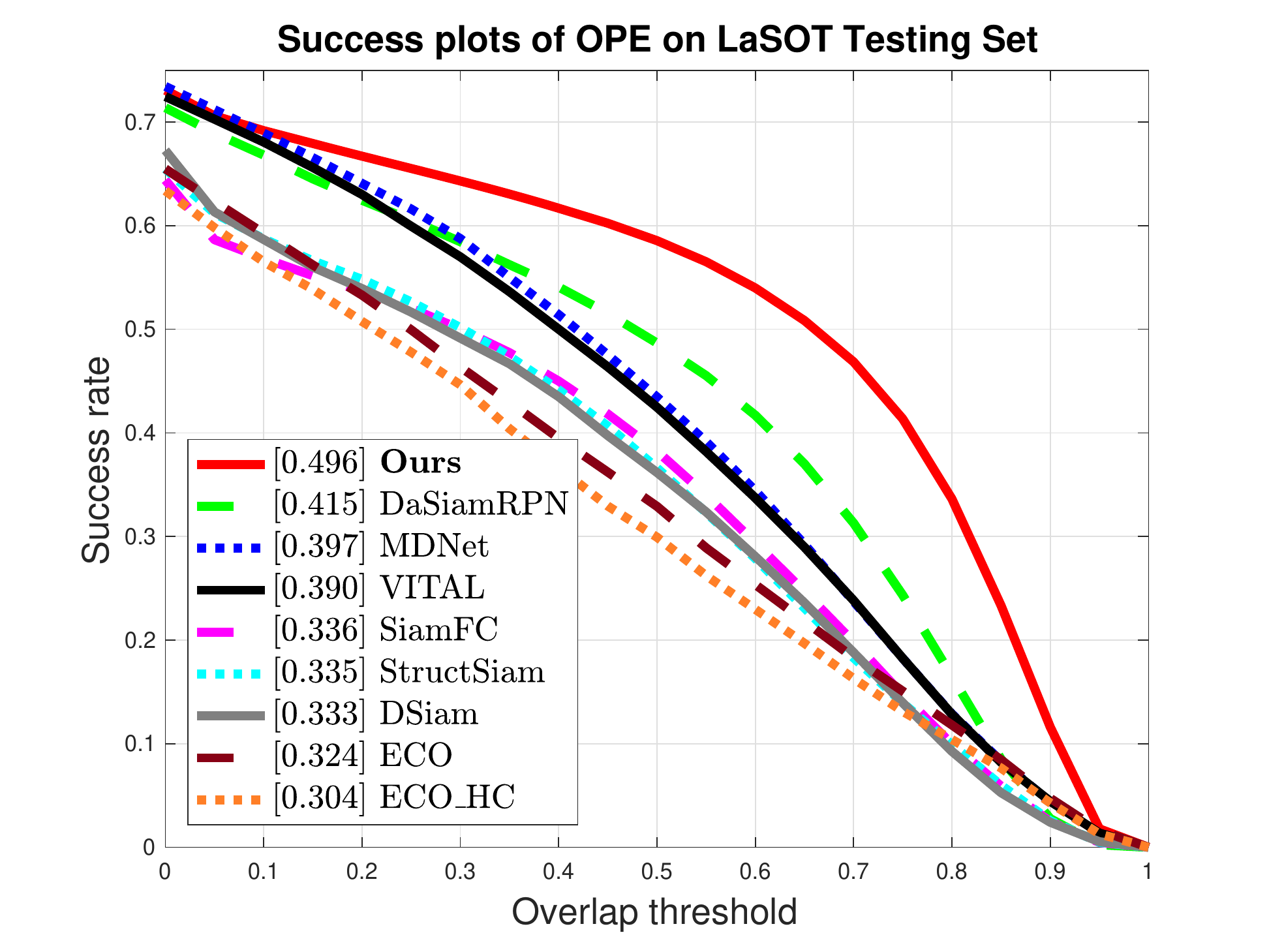}}
\subfigure{\includegraphics[trim={1cm 0cm 1cm 0cm}, clip, width=0.495\linewidth]{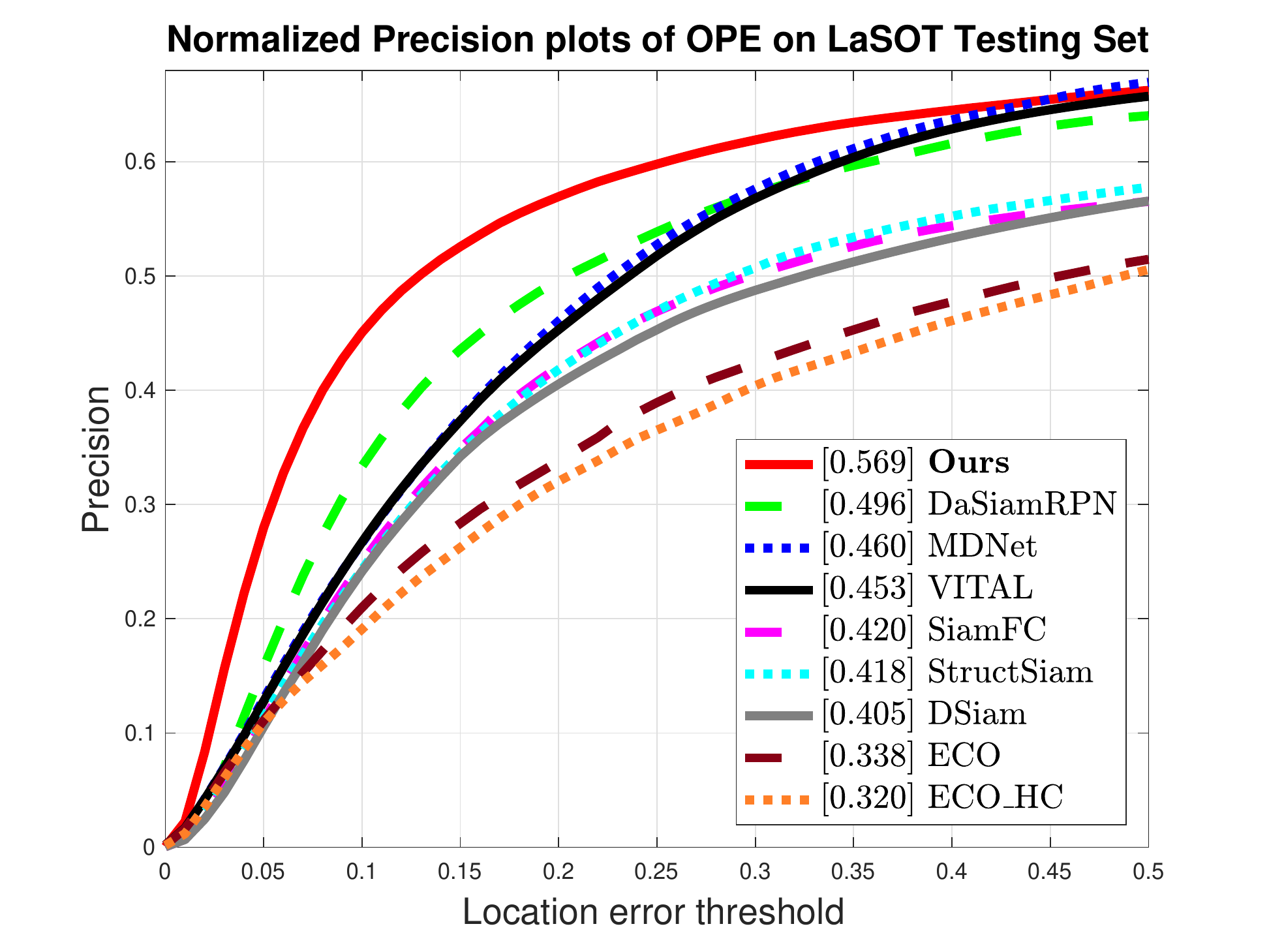}}
\end{center}
\vspace{-5mm}
\caption{Evaluation results of trackers on LaSOT.}
\vspace{-4mm}
\label{fig:lasot}
\end{figure}

\vspace{-4mm}
\paragraph{LaSOT Dataset.}
To further validate the proposed framework on a larger and more challenging dataset, we conduct experiments on LaSOT~\cite{ARX18LaSOT}. The LaSOT dataset provides a large-scale, high-quality dense annotations with 1,400 videos in total and 280 videos in the testing set. Fig.~\ref{fig:lasot} reports the overall performances of our SiamRPN++ tracker on LaSOT testing set. Without bells and whistles, our SiamRPN++ model is sufficient to achieve state-of-the-art AUC score of $49.6\%$. Specifically, SiamRPN++ increases the normalized distance precision and AUC relatively by $23.7\%$ and $24.9\%$ over MDNet \cite{CVPR16MDNet}, which is the best tracker reported in the original paper.

\newcommand{\demph}[1]{\textcolor{demphcolor}{#1}}
\renewcommand\arraystretch{1.1}
\setlength{\tabcolsep}{1.pt}
\begin{table}[t]
\centering
\footnotesize
\begin{tabular}{c|cccccccc}
& \begin{tabular}[c]{@{}l@{}}CSRDCF\\ [-.7ex]~~~~\cite{CVPR17CSRDCF}\end{tabular} 
& \begin{tabular}[c]{@{}l@{}}ECO\\ [-.7ex] ~~\cite{CVPR17ECO}\end{tabular} 
& \begin{tabular}[c]{@{}l@{}}SiamFC\\ [-.7ex] ~~~~\cite{ECCV16SiamFC}\end{tabular} 
& \begin{tabular}[c]{@{}l@{}}CFNet\\ [-.7ex] ~~~~\cite{CVPR17CFNet}\end{tabular}
& \begin{tabular}[c]{@{}l@{}}MDNet\\ [-.7ex]~~~\cite{CVPR16MDNet}\end{tabular}
& \begin{tabular}[c]{@{}l@{}}DaSiamRPN\\ [-.7ex]~~~~~~~\cite{ECCV18DaSiamRPN}\end{tabular}
& \begin{tabular}[c]{@{}l@{}}\textbf{Ours}\\ [-.7ex] {}\end{tabular}\\
\shline
AUC ($\%$)     & 53.4 & 55.4 & 57.1 & 57.8 & 60.6 & 63.8 & \color{red}\textbf{73.3} \\[-.2ex]
P ($\%$)   & 48.0 & 49.2 & 53.3 & 53.3 & 56.5 & 59.1 & \color{red}\textbf{69.4} \\[-.2ex]
P$_{norm}$ ($\%$) & 62.2 & 61.8 & 66.3 & 65.4 & 70.5 & 73.3 & \color{red}\textbf{80.0} \\[-.2ex]
\end{tabular}
\vspace{.5em}
\caption{State-of-the-art comparison on the TrackingNet \texttt{test} set in terms of success, precision, and normalized precision.}
\label{tab:trackingnet}
\vspace{-1em}
\end{table}

\vspace{-4mm}
\paragraph{TrackingNet Dataset.}
The recently released TrackingNet~\cite{ECCV18trackingnet} provides a large amount of data to assess trackers in the wild. We evaluate SiamRPN++ on its \texttt{test} set with 511 videos. Following~\cite{ECCV18trackingnet}, we use three metrics success (AUC), precision (P) and normalized precision (P$_{norm}$) for evaluation. Table~\ref{tab:trackingnet} demonstrates the comparison results to trackers with top AUC scores, showing that SiamRPN++ achieves the best results on all three metrics. In specific, SiamRPN++ obtains the AUC score of $73.3\%$, P score of $69.4\%$ and P$_{norm}$ score of $80.0\%$, outperforming the second best tracker DaSiamRPN~\cite{ECCV18DaSiamRPN} with AUC score of $63.8\%$, P score of $59.1\%$ and P$_{norm}$ score of $73.4\%$ by $9.5\%$, $10.3\%$ and $6.6\%$, respectively.

In summary, it is important to note that all these consistent results show the generalization ability of SiamRPN++.

\section{Conclusions}
\label{sec_conclusion}
In this paper, we have presented a unified framework, referred as SiamRPN++, to end-to-end train a deep Siamese network for visual tracking. 
We show theoretical and empirical evidence that how to train a deep network on Siamese tracker.
Our network is composed of a multi-layer aggregation module which assembles the hierarchy of connections to aggregate different levels of representation and a depth-wise correlation layer which allows our network to reduce computation cost and redundant parameters while also leading to better convergence. 
Using SiamRPN++, we obtained state-of-the-art results on the VOT2018 in real-time, showing the effectiveness of SiamRPN++. SiamRPN++ also acheived state-of-the-art results on large datasets like LaSOT and TrackingNet showing its generalizability.

{\small
\bibliographystyle{ieee}
\bibliography{SiamTrack}
}

\end{document}